\documentclass[10pt, a4paper]{article}

\usepackage[]{lrec-coling2024} 

\usepackage{url}
\usepackage{latexsym}
\usepackage{graphicx}
\usepackage{color}
\usepackage{xcolor}
\usepackage{tikz}

\newcommand\rev[1]{#1}
\usetikzlibrary{arrows}
\usepackage{colonequals}
\usepackage{booktabs}
\usepackage{subcaption}
\usepackage{multirow}
\usepackage{amsmath}
\usepackage{placeins}
\usepackage{amssymb}
\usepackage[normalem]{ulem}

\newtheorem{prop}{Proposition}
\newtheorem{corr}{Corollary}
\newtheorem{algo}{Algorithm}
\usepackage{physics}
\DeclareMathOperator*{\argmax}{arg\,max}
\DeclareMathOperator*{\argmin}{arg\,min}

\title{On a Novel Application of Wasserstein-Procrustes \\ for Unsupervised Cross-Lingual Alignment of Embeddings}

\name{Guillem~Ram\'irez$^*$$^1$\thanks{$^*$Equal contribution.}, Rumen~Dangovski$^*$$^1$, Preslav~Nakov$^2$, Marin~Solja\v{c}i\'c$^1$} 

\address{Massachusetts Institute of Technology (MIT)$^1$\\ Mohamed bin Zayed University of Artificial Intelligence (MBZUAI)$^2$ \\
         gramirez@ed.ac.uk}

\abstract{
Unsupervised word embeddings, pre-trained on vast monolingual text corpora, have driven the neural revolution in Natural Language Processing (NLP). Initially developed for English, these embeddings soon expanded to other languages, spurring efforts to align embedding spaces for cross-lingual NLP applications. Unsupervised cross-lingual alignment of embeddings (UCAE) is particularly appealing due to its minimal data requirements and competitive performance against supervised and semi-supervised approaches. In this work, we scrutinize prevalent UCAE methods and discover their objectives inherently resemble the Wasserstein-Procrustes problem. Consequently, we propose a direct solution for Wasserstein-Procrustes, enhancing popular UCAE techniques such as iterative closest point (ICP), multilingual unsupervised and supervised embeddings (MUSE), and supervised Procrustes methods. Evaluation on benchmark datasets demonstrates significant improvements over existing approaches. Our reexamination of the Wasserstein-Procrustes problem fosters further research, paving the way for more effective algorithms to align word embeddings across languages. 
 \\ \newline  }

\begin{document}

\maketitleabstract

\section{Introduction}
\label{introduction}
Pre-trained word embeddings, which map words to dense vectors of low dimensionality, have been the key enabler of the ongoing neural revolution, and today they serve as the basic building blocks of contemporary Natural Language Processing (NLP) models. While initially introduced for English \citep{mikolov2013efficient, pennington2014, bojanowski2017enriching, joulin2017bag}, pre-trained embeddings quickly emerged for a number of other languages \citep{heinzerling2017bpemb}, and the idea of cross-language embedding spaces was born. In a cross-language embedding space, two semantically similar (or dissimilar) words would be close to (or far from) each other regardless of whether they are from the same or from different languages. Using such a space is attractive, as for a number of NLP tasks, it enables the application of an NLP model trained for one language on input from another language.

Ideally, such spaces could be trained on parallel bilingual datasets, but such resources are of limited size, e.g.,~compared to the large-scale monolingual resources typically used to pre-train monolingual word embeddings.
Thus, it has been more attractive to train monolingual word embeddings for different languages independently, and then to try to align the corresponding embedding spaces in what is commonly known as bilingual lexicon induction. This has been attempted in a supervised~\citep{DBLP:journals/corr/MikolovLS13, inproceedings-irvine}, in a semi-supervised~\citep{artetxe-etal-2017-learning}, and in an unsupervised setting~(\citet{lample2017unsupervised, lample2019,alipour2022learning,feng2022cross,tian2022rapo,liang2023xlm,li2023dual,liu2023topic,ghayoomi2023training,ghazvininejad2023dictionary}).

Initial space alignment efforts used word translation pairs as anchors, inferring transformations between languages in a supervised setup \citep{DBLP:journals/corr/MikolovLS13}. The alignment employs an orthogonal transformation minimizing the Frobenius norm in the Procrustes problem, with a closed-form solution obtainable via SVD.
For the translation of word embeddings, $W$ is taken to be an orthogonal matrix due to a self-similarity argument \citep{smith2017offline}. The convenience of using an orthogonal matrix has also been supported empirically \citep{xing-etal-2015-normalized, zhang-etal-2016-ten, artetxe-etal-2016-learning}. The orthogonal Procrustes problem has a closed-form solution $W = UV^{\top}$, where $U\Sigma V^{\top}$ is the singular value decomposition (SVD) of $X^{\top} Y$ as shown by~\citet{Schonemann1966}. 

\paragraph{Procrustes} Given two ordered clouds of points $X$, $Y  \in  \mathbb{R}^{N \times d}$, each with $N$ points of dimension $d$, the orthogonal Procrustes problem finds the orthogonal matrix $W \in  \mathbb{R}^{d\times d}$ that minimizes the following Frobenius norm: 
\begin{equation}
\argmin_{W\in O(d)}\norm{XW - Y}_{2}^{2}
\end{equation}

A popular unsupervised formulation of the problem is known as the Wasserstein-Procrustes \citep{DBLP:conf/aistats/GraveJB19, alaux2018unsupervised}, which is more challenging as it needs to optimize a generalization of the Procrustes objective. One-to-one maps are encouraged through a permutation matrix $P$. 

The convenience of one-to-one maps is justified for different reasons. First, the hubness problem \citep{dinu2014improving} occurs in high-dimensional vector spaces where certain vectors are the nearest neighbor to a disproportionate number of other vectors, thus reducing the
quality of the embedding space \citep{JMLR:v11:radovanovic10a}. Second, one-to-one maps can be linked to Wasserstein distance and computational optimal transport.
\paragraph{Wasserstein-Procrustes} Given two clouds of points $X$, $Y  \in  \mathbb{R}^{N \times d}$, each with $N$ points of dimension $d$, the Wasserstein-Procrustes problem finds an orthogonal matrix $W \in  \mathbb{R}^{d\times d}$ and a permutation matrix $P \in  \mathbb{R}^{N\times N}$ that minimize the Frobenius norm:
\begin{equation}
   \argmin_{P\in \pi(N), W\in O(d)} \norm{XW - PY }_{2}^{2}
 \label{WP}
\end{equation}
where $\pi(N)$ is the set of $N$-dimensional permutation matrices and $O(d)$ is the set of $d$-dimensional orthogonal matrices.

In practice, most approaches modify the objective yet achieve good accuracy in synthetic dictionary induction tasks. We ask: Can we find approximate Wasserstein-Procrustes solutions (Equation~\ref{WP}) with high accuracy in dictionary tasks? Can we enhance existing methods using refinements to optimize Equation~\ref{WP}? Can we identify scenarios with good solutions? We address these questions by analyzing different objective functions in the literature, adhering to \citet{artetxe2020rigor}'s call for fair model comparison.

\section{Background: Towards a Unifying Framework}
\label{background}

There have been attempts to compare different methods proposed for \rev{the Unsupervised Cross-Lingual Alignment of Embeddings, or UCAE} \citep{NIPS2019_8836}, and there have been papers that have tried to generalise the different possibilities one approach could possibly have. \citet{Artetxe2018GeneralizingAI} proposed a framework based on different steps and showed how existent methods would fit in it. \citet{Ruder_2019} described the most general framework for \rev{UCAE}. 
However, we are not aware of a unified description of the existing methods from the point of view of what is being optimized, namely the loss function. 
We start by analyzing methods based on optimal transport methods, as they are most relevant to our approach. 

\subsection{Optimal Transport Methods}
There have been some approaches framing the problem of unsupervised dictionary induction as an optimal transport problem, and this is the approach we will adopt in the following sections. \citet{haghighi-etal-2008-learning} proposed a self-learning method for bilingual lexicon induction, representing words with orthographic and contextual features and using the Hungarian algorithm~\citep{Tomizawa1971OnST} to find an optimal one-to-one matching.

With the emergence of word embeddings~\citep{mikolov2013efficient}, words were interpreted as vectors in high-dimensional spaces, and concepts such as distance between words started to gain attention. \citet{ruder-etal-2018-discriminative} presented Viterbi EM, where words were mapped following a one-to-one map between subsets $X'$ and $Y'$ of $X$ and $Y$, respectively, and the isometry was induced by an orthogonal matrix. They deviated from the Wasserstein-Procrustes objective by including a penalization term for unmatched words $Y_\perp'=Y-Y'$. They did not consider all possible matches, instead imposing a restriction on the $k$ nearest neighbors when running the Jonker-Volgenant algorithm for optimal transport \citep{10.1007/BF02278710}. 

\citet{zhang-etal-2017-earth} proposed two different methods: WGAN (an adversarial network that optimizes the Wasserstein distance) and EMDOT (an iterative procedure that uses Procrustes and solves a linear transport problem). Both methods are inspired by the Earth Mover’s Distance (EMD), which defines a distance between probability distributions, which they applied to frequencies of words. They found that, although EMDOT could converge to bad local minima, it improved the results when used as a refinement tool after first optimizing with WGAN.
\citet{alvarez-melis-jaakkola-2018-gromov} used the concept of Gromov-Wasserstein distance to provide an alternative to Wasserstein-Procrustes. This distance does not operate on points but on pairs of points, turning the problem of finding optimal matching $\Gamma^*$ from a linear into a quadratic one. This new loss function can be optimized efficiently with first-order methods, whereby each iteration involves solving a traditional optimal transport problem. \citet{artetxe-etal-2018-robust} achieved better results by combining this idea with a refinement method called stochastic dictionary induction, i.e.,~randomly dropping dimensions out of the similarity matrix when extracting a seed dictionary for the next iteration of the Procrustes analysis.

\subsection{Other Methods}

Wasserstein-Procrustes is one of the recurring loss functions in the literature, but there have been also deviations from the original problem. \citet{DBLP:conf/aistats/GraveJB19} suggested an iterative procedure whose initial condition minimizes the convex relaxation $\norm{X^{\top} PY}^{2}_{2}$ instead of the original problem. This relaxation is known as the Gold-Rangarajan relaxation and can be solved using the Frank-Wolfe algorithm \citep{491619, doi:10.1002/nav.3800030109}. The solution to this relaxation is then used as the initial condition for a gradient-based iterative procedure that stochastically samples different subsets of words for which there is not necessarily a direct translation. 

This deviates strongly from Objective~\ref{WP}: not only the initial condition does not optimize the Wasserstein-Procrustes objective, but also the iterative procedure does not optimize it, as it translates words that are not necessarily the optimal matches. \citet{alaux2018unsupervised} were also inspired by Objective~\ref{WP} for aligning multiple languages in a common vector space. However, they 
minimized a loss function based on the CSLS metric from \citet{conneau2017word}. In a similar fashion, the entropy regularization of the Gromov-Wasserstein problem \citep{article-memoli} has been used for bilingual lexicon induction.

Generative Adversarial Network (GAN) optimization was first introduced for bilingual lexicon induction by \citet{barone-2016-towards}, but its canonical implementation was given by \citet{conneau2017word}, who presented \emph{multilingual unsupervised and supervised embeddings} (MUSE), an adversarial method in which the transformation matrix $W$ is considered as a generator, and thus is trained by a generative adversarial network, so that the mapped word embeddings $XW$ cannot be distinguished from the set $Y$ via a discriminator~\citep{10.5555/2969033.2969125}. However, a simple thought experiment can convince us that this approach does not minimize distances.
We elaborate on that experiment in the Appendix.

\citet{hoshen2018nonadversarial} were inspired by the Iterative Closest Point (ICP) method used in 3D point cloud alignment. Although their transformation matrix is not necessarily orthogonal, this property is enforced using the regularization $L(X,Y,W;\lambda)\colonequals \lambda \|XWW^{\top}-X\|_2^2 + \lambda \|YW^{\top} W - Y\|_2^2$. 
Another fundamental difference to Objective~\ref{WP} is that they do not use a one-to-one mapping for $P$.

This list is not exhaustive, as there have been successful methods that do not rely on loss functions, and such that go beyond
the geometry of the trained word embeddings. For example, \citet{artetxe2019acl-bilingual} used both the word embeddings and the monolingual corpus used to train them.

To sum up, in Table~\ref{formalism}, we list the relevant objectives from above using our formalism from Equation~\ref{WP}. In the table, $\Gamma^*$ is the optimal Gromov-Wasserstein matching, $X'$ and $Y'$ are subsets of the corresponding $X$ and $Y$, $Y_\perp'$ is the complement of $Y'$ in $Y$, and $\overline{Y_\perp'}$ is the average of the complements.

\begin{table*}[ht]
  \centering
  \small
  \begin{tabular}{lc@{}}
    \toprule
    \multicolumn{1}{c}{\bf Method} & \multicolumn{1}{c}{\bf Objective} \\
    \cmidrule(r){1-1}
    \cmidrule(r){2-2}

    \citet{DBLP:conf/aistats/GraveJB19} and Ours & $\min_{W \in O(d), P \in \pi(N)} \norm{XW - PY }_{2}^{2}$ \\
    \citet{alvarez-melis-jaakkola-2018-gromov} & $\min_{\Gamma^* \text{ best coupling}, W \in O(N)} \|X\Gamma^* - WY\|^2_2$ \\
    \citet{hoshen2018nonadversarial}  & $\min_{W \in O(d)}\|XW-Y\|_2^2+\|YW^{\top}-X\|_2^2 + L(X,Y,W;\lambda)$ \\ 
    \citet{ruder-etal-2018-discriminative} & $\min_{W \in O(d), P' \in \pi(N')} \|X'W-P'Y'\|^2_2 + \left \|Y'_\perp-\overline{Y'_\perp}\right \|^2_2$ \\
    \citet{lample2017unsupervised} & $\min_{W} \max_{\theta_D}  \mathbb{P}_{\theta_D}(\text{source}|WX)\mathbb{P}_{\theta_D}(\text{target}|Y)$ \\
    \citet{zhang-etal-2017-earth} & $\min_{W \in O(d), P \in 
    \pi(N)} \sum_{i=1,j=1}^{N,N} P_{i,j}\left(\left(X_iW\right)_j-Y_i\right)^2$ \\
    \bottomrule
  \end{tabular}
  \caption{Objective functions of relevant existing methods in the language of our formalism.
  }
  \label{formalism}
\end{table*}

\section{Properties of the Wasserstein-Procrustes Problem}
\label{sec:properties}

We begin by simplifying Objective~\ref{WP} to arrive at some essential properties, described below.

\begin{prop}[\citet{DBLP:conf/aistats/GraveJB19}]
\label{prop:main}
The Wasserstein-Procrustes problem is equivalent to maximizing the trace norm on the permutation matrix $X^{\top} PY$ over $P$, described as follows:

\begin{equation}\label{trace_noorm}
\argmin_{P\in \pi(N), W\in O(d)}   \norm{XW - PY}_{2}^{2}=\argmax_{P\in \pi(N)} \norm{X^{\top} PY}_{*}
\end{equation}
where $\norm{\cdot}_{\**}$ denotes the nuclear norm and $W$ is selected, so that it fulfills that $U^{\top}  W V = \mathbb{I}_d$, where both $U(P)$ and $V(P)$ are evaluated at a matrix $P^{*}$ that achieves the optimum of Equation~\ref{trace_noorm}.
\end{prop}


\paragraph{Hungarian algorithm} Given two clouds of points $X$, $Y  \in  \mathbb{R}^{N \times d}$, each with $N$ points of $d$ dimensions, the Hungarian algorithm finds the permutation matrix $P$ that gives the correspondence between the different points by solving the following problem:
\begin{equation}
\label{hungares}
\argmin_{P\in \pi(N)}\norm{X - PY}_{2}^{2}.
\end{equation}

\rev{Replacing $W$ in Proposition~\ref{prop:main} with the identity matrix $\mathbb{I}_d$ and noting that $\langle \mathbb{I}_d, X^\top PY \rangle_2 = \Tr\left(X^{\top} PY\right)$ holds for the Frobenius inner product, we obtain the following:}

\begin{corr}
\label{prop:hung}

Problem~\ref{hungares} is equivalent to maximizing the trace of $X^{\top} PY$ over $P$:
\begin{equation}
\argmin_{P\in \pi(N)}\norm{X - PY}_{2}^{2} = \argmax_{P\in \pi(N)}\Tr\left(X^{\top} PY\right),
\label{trace_hung}
\end{equation}
which is the maximum weight matching problem. The latter can be solved using the Hungarian algorithm, which has a complexity of $O(N^3)$ \citep{Tomizawa1971OnST}. 
\end{corr}

\rev{Even though the Hungarian algorithm has cubic complexity, we could still run it feasibly for $N=45,000$. In principle, our refinement methods work well by using a subset of the full vocabulary, which typically has $N=200,000$ words. Speedups of the Hungarian algorithm and approximations could be pursued in future work.}

\paragraph{Equivalent problems} One useful property of the trace norm is that $\norm{UA}_{\**} = \norm{AV}_{\**} = \norm{A}_{\**}$, where $U$ and $V$ are orthogonal matrices. Knowing this, and writing $U_{X}\Sigma_{X}V_{X}^{\top}$ and  $U_{Y}\Sigma_{Y}V_{Y}^{\top}$ as the SVD decompositions for $X$ and $Y$, respectively, we obtain the following: 

\begin{equation}
 \norm{X^{\top} PY}_{\**} = \norm{V_{X}\Sigma_{X}U_{X}^{\top} PU_{Y}\Sigma_{Y}V_{Y}^{\top} }_{\**} 
\end{equation}

which yields
\begin{equation} \label{trace-norm}
\argmax_{P\in \pi(N)} \norm{\Sigma_{X}U_{X}^{\top} PU_{Y}\Sigma_{Y}}_{\**}. 
\end{equation}

Let us define $\widetilde{X} = U_{X}\Sigma_{X}$ and $\widetilde{Y} = U_{Y}\Sigma_{Y}$. Then, the optimal solution $P$ would be the same for translations involving all of the following pairs of word embeddings: ($X$, $Y$), ($\widetilde{X}$, $Y$), ($X$, $\widetilde{Y}$) and ($\widetilde{X}$, $\widetilde{Y}$). However, the optimal transformation matrix $W^{\**}$ will be different for each of these problems. There is a different, yet interesting way of looking at this: if we follow the iterative procedure that starts from an initial transformation matrix $X_{0} = X W_{0}$ (where $W_{0}$ is our initial approximation to the transformation matrix), and then we want to solve Problem~\eqref{trace_hung}, the equivalent problems will induce a set of \textit{natural initializations} of the transformation $W$, which we formalize below: 
\begin{quote}
\emph{Given the iterative procedure that tries to minimize the Wasserstein-Procrustes objective by first obtaining the permutation matrix $P_{n} = \argmin_{P\in \pi(N)}\mathrm{Tr}(X_{n}^\top PY_{n})$ and then the transformation matrix $W_{n}  = \argmin_{W\in \mathbb{R}^{N \times N}}\norm{X_{n}W - P_{n}Y_{n}}_{2}^{2}$, the procedure aims for the same solution $P$ as the problems with initial conditions $X_{0} = X W_{0}$, $X_{0} = X V_{X} W_{0}$, $X_{0} = X W_{0} V_{Y}^{\top} $, $X_{0} = X  V_{X} W_{0} V_{Y}^{\top} $.}
\end{quote}

The significance of the different natural initialization is that it gives us a starting point for different problems that have the same solution $P$. It must be noted, however, that these transformations of $X_{0}$ are not the unique ones that will have the same original solution, as the trace norm is invariant to any orthogonal transformation; however, they help to avoid bad local minima as we will show in Section~\ref{sec:experiments} below. Another way of looking at these initialization is that we are performing PCA to the embedding matrices  without a dimensionality reduction. \citet{hoshen2018nonadversarial} proposed using PCA in a similar context.

\section{Approach}
 \label{sec:approach}
Below, we present a general iterative algorithm to solve the Wasserstein-Procrustes problem. 

\paragraph{Joint optimization on $W$ and $P$.} For the Wasserstein-Procrustes problem from Equation \ref{WP}, a joint iterative procedure involving the Procrustes problem and the Hungarian algorithm (see Algorithm ~\ref{CIH}) has been dismissed due to its computational cost and convergence to bad local minima~\cite{zhang-etal-2017-earth}. However, as we will show below, there are a number of situations where such an approach can be extremely beneficial if we apply some improvements based on the discussion in the previous section.
\begin{algo}
\label{CIH}
\textit{Cut Iterative Hungarian (CIH) Algorithm}
\begin{enumerate}
\item We initialize as follows: $X \leftarrow X W_{0}$.
\item We find $P \leftarrow \text{Hungarian}\left(X, Y\right)$ and $W \leftarrow \text{Procrustes}\left(X, PY\right).$
\item If the trace norm has increased, update $X_{NEW} \leftarrow XW$ and $Y_{NEW} \leftarrow PY$, repeat Step 2. 
\end{enumerate}
\end{algo}

\paragraph{Variants of the natural initializations.} The first improvement is to consider the different equivalent problems or the natural initialization transformations, mentioned in the previous section. We observe empirically that apart from the four problems that share the same optimal $P$, it is possible to improve the results by considering the opposite optimization problem: instead of maximizing the costs for the two clouds of points $(X, Y)$, sometimes \emph{minimizing} the costs yields a solution with a higher trace norm, and thus the algorithm eventually converges to a better solution. \rev{The matrix $X^\top PY$ is generally not symmetric with non-negative eigenvalues, and thus the trace norm and the trace are not the same.}
The minimization is achieved by simply considering the cloud $-X$ instead of $X$. Algorithm~\ref{IH} is the most general iterative procedure that we consider here, and it serves as the backbone for our experiments below:

\begin{algo}
\label{IH}
\textit{Iterative Hungarian (IH) Algorithm}. It is the same as Algorithm~\ref{CIH}, but in Step 2 we also consider the solutions for four \textit{natural} initializations: $X_{0} = X W_{0}$, $X_{0} = X V_{X} W_{0}$, $X_{0} = X W_{0} V_{Y}^{\top} $, $X_{0} = X  V_{X} W_{0} V_{Y}^{\top} $, also considering the cloud $-X$ for the four different initializations.  
\end{algo}

\paragraph{Supervised translation.} Although the scope of this paper is the unsupervised cross-lingual alignment of embeddings, we also decided to run some experiments that involve minimal supervision. There are different ways of doing this, but the procedure that converges the fastest is to fix $n$ pairs of words when calculating the Hungarian map, where typically $n \ll N$. We also consider similar approaches, e.g.,~deciding how to update Algorithm~\ref{IH}, taking into account the accuracy of the maps on a small subset of the data. Choosing among these methods could be motivated by how trustworthy the initial dictionary is. By \emph{trustworthy} here we mean how many of the corresponding cloud points are correctly matched.

We use a fast implementation of the Hungarian algorithm\footnote{\url{http://github.com/cheind/py-lapsolver}} for dense matrices based on shortest path augmentation \citep{10.1145/321694.321699}. Relaxations of the original problem can achieve higher speed ups. \citet{cuturi2013sinkhorn} showed how smoothing the classical optimal transport problem with an entropic regularization term results in a problem that can be solved using the Sinkhorn-Knopp's matrix scaling algorithm~\cite{sinkhorn1967} at a speed that is orders of magnitude faster than that of transportation solvers.

\paragraph{Mapping.} Although our method finds a permutation matrix $P$, this is not necessarily the best possible mapping as the set of word-to-word translations does not have to represent a one-to-one mapping. Nearest neighbor approaches can be used, but they suffer from the so-called hubness problem: in high-dimensional vector spaces, certain vectors are universal nearest neighbors \citep{JMLR:v11:radovanovic10a}, and this is a common problem for word-embedding-based bilingual lexicon induction \citep{dinu2014improving}. \citet{conneau2017word} presented \textit{cross-domain similarity local scaling} (CSLS), which is a method intended to reduce the influence of hubs by expanding high-density areas and condensing low-density ones. 

Given a source vector $x_s$, the mean similarity of its transformation $Wx_s$ to its $k$ target nearest neighbors $\mathcal{N}_T^k(Wx_s)$ is defined as $$\mu^{k}_{T}(Wx_s) = \frac{1}{k} \sum_{y_t \in \mathcal{N}_T^k(Wx_s)} \cos{(Wx_s, y_t)}.$$
Likewise is defined $\mu^{k}_{S}(y_t)$, i.e.,~the mean similarity of a target word $y_t$ to its neighborhood of source mapped vectors. Then, the CSLS similarity between a mapped source vector $x_s$ and a target vector $y_t$ is calculated as follows: $\mathrm{CSLS}(Wx_s, y_t) = 2\cos{(Wx_s, y_t)} - \mu^{k}_{T}(Wx_s) - \mu^{k}_{S}(y_t)$. Intuitively, this mapping increases the similarity associated with isolated word vectors, and it decreases the one for vectors lying in dense areas. In the following experiments, we use the mapping induced by CSLS with $k = 10$.


\section{Experiments}
 \label{sec:experiments}
Below, we describe our experiments. In our first set of experiments, we deploy our method on top of well-known methods for cross-lingual alignment of embeddings and we show that it improves their accuracy, meaning that it can be used as a refinement tool. In the second set of experiments, we recreate the benchmarks from \citep{DBLP:conf/aistats/GraveJB19}, and we show that our method can align word embedding spaces without a good initialization matrix. 

\subsection{The Iterative Hungarian Algorithm as a Refinement Tool} 
The experiments in this section use the Iterative Hungarian (IH) algorithm starting with the initial condition $W_{0}$ produced from the following methods: 
\begin{itemize}
    \item The adversarial approach by \citet{lample2017unsupervised}. This combines the adversarial training described in Section~\ref{background} with a refinement step, which consists of creating a dictionary from the best matches and then running the supervised Procrustes algorithm using that dictionary.
    \item The supervised Procrustes approach.
    \item The Iterative Closest Point (ICP) method by \citet{hoshen2018nonadversarial}.
\end{itemize}

We used the word embeddings, the dictionaries and the evaluation methods from \citet{conneau2017word}. We trained the transformation matrix obtained from MUSE \cite{conneau2017word} on 200,000 words. Then we ran the Iterative Hungarian algorithm on a subsample of 45,000 words. Finally, we refined the new transformation matrix following the procedure in \citet{conneau2017word}. Also, inspired by their work, we induced mappings using CSLS with $k=10$ nearest neighbors.  

\begin{table*}[ht]
  \centering
  \begin{tabular}{llllllllllll}
    \toprule
    \multicolumn{1}{c}{Method} & \multicolumn{1}{c}{en-es} & \multicolumn{1}{c}{es-en} & \multicolumn{1}{c}{en-fr} & \multicolumn{1}{c}{fr-en}  & \multicolumn{1}{c}{en-it} & \multicolumn{1}{c}{it-en}  & \multicolumn{1}{c}{en-de} & \multicolumn{1}{c}{de-en} & \multicolumn{1}{c}{en-ru} & \multicolumn{1}{c}{ru-en} & \multicolumn{1}{c}{mean} \\
    \cmidrule(r){1-1}
    \cmidrule(r){2-2}
    \cmidrule(r){3-3}
    \cmidrule(r){4-4}
    \cmidrule(r){5-5}
    \cmidrule(r){6-6}
    \cmidrule(r){7-7}
    \cmidrule(r){8-8}
    \cmidrule(r){9-9}
    \cmidrule(r){10-10}
    \cmidrule(r){11-11}
    \cmidrule(r){12-12}
  MUSE (1)  & \textbf{82.6}                                              & 83.7                                                       & 82.5                                                       & 82.0                                                         & 76.8                                                       & 77.6                                                       & \textbf{75.1}                                              & 72.5                                                       & 42.5                                             & 60.1                                                       & 73.5                                \\
   MUSE (1) + IH & 82.5                                             & \textbf{84.1}                                              & \textbf{82.7}                                              & \textbf{82.4}                                              & \textbf{78.3}                                              & \textbf{77.9}                                              & 74.9                                             & \textbf{73.3}                                              & \textbf{44.5}                                              & \textbf{60.7}                                              & \textbf{74.1}                       \\  
    \cmidrule(r){1-12}
MUSE (2) & 81.9                                                       & 83.2                                                       & 82.1                                                       & 82.4                                                       & \textbf{77.5}                                              & 77.5                                                       & 74.7                                                       & 72.9                                                       & 37.0                                                & 61.9          & 73.1                                \\

MUSE (2) + IH & \textbf{82.5}                                              & \textbf{84.1}                                              & \textbf{82.7}                                              & 82.4                                              & 77.3                                             & \textbf{78.1}                                              & 74.7                                            & \textbf{73.3}                                              & \textbf{42.3}                                              & \textbf{62.5}                                              & \textbf{74.0}                       \\
\cmidrule(r){1-12}

MUSE (3) & 82.1                                                       & \textbf{84.0}                                                & 82.1                                                       & 82.3                                                       & \textbf{77.9}                                              & 77.7                                                       & 74.8                                                       & 69.9                                                       & 37.1                                                       & 60.1                                                       & 72.8                                 \\

MUSE (3) + IH & \textbf{82.3}                                              & 83.9                                              & \textbf{82.6}                                              & \textbf{82.4}                                              & 77.8                                              & \textbf{77.8}                                              & \textbf{75.1}                                              & \textbf{72.9}                                              & \textbf{38.9}                                              & \textbf{62.1}                                              & \textbf{73.6} \\
\cmidrule(r){1-12}
\cmidrule(r){1-12}
 Procrustes      & 81.7                        & 83.3                        & 82.1                        & 81.9                        & 77.3                        & 77.0                        & 73.7                        & 72.7                        & \textbf{49.9}               & 60.8                        & 74.0                     \\
Procrustes + IH & \textbf{82.5}               & \textbf{84.2}               & \textbf{82.2}               & \textbf{82.6}               & \textbf{78.1}               & \textbf{78.0}               & \textbf{75.0}               & \textbf{73.5}               & 47.9                        & \textbf{63.9}               & \textbf{74.8}            \\
\cmidrule(r){1-12}
\cmidrule(r){1-12}
ICP (1)         & 81.9                        & 82.7                        & 81.9                        & 81.5                        & 76.0                        & 75.5                        & 72.3                        & 72.3                        & \textbf{46.4}               & 56.6                        & 72.7                     \\
ICP (1) + IH    & \textbf{82.5}               & \textbf{84.1}               & \textbf{82.1}               & \textbf{82.7}               & \textbf{78.1}               & \textbf{78.0}               & \textbf{76.6}               & \textbf{72.7}               & 46.2                        & \textbf{63.2}               & \textbf{74.6}            \\ 
\cmidrule(r){1-12}
ICP (2)        & 80.8                        & 82.5                        & 81.3                        & 80.4                        & 76.3                        & 76.3                        & 72.3                        & 72.4                        & 46.5                        & 57.5                        & 72.6                     \\
ICP (2) + IH    & \textbf{82.2}               & \textbf{84.1}               & \textbf{82.4}               & \textbf{82.3}               & \textbf{78.2}               & \textbf{77.9}               & \textbf{76.4}               & \textbf{73.3}               & \textbf{46.6}               & \textbf{63.1}               & \textbf{74.7}            \\ 
\cmidrule(r){1-12}
ICP (3)        & 82.0                        & 82.6                        & 82.0                        & 81.8                        & 75.7                        & 76.6                        & 73.1                        & 72.6                        & 45.1                        & 56.2                        & 72.8                     \\
ICP (3) + IH    & \textbf{82.5}               & \textbf{84.2}               & 82.0                        & \textbf{82.4}               & \textbf{77.7}               & \textbf{77.7}               & \textbf{76.9}               & \textbf{73.5}               & \textbf{45.2}               & \textbf{63.1}               & \textbf{74.5}            \\

  \bottomrule
  \end{tabular}
  \caption{The Iterative Hungarian (IH) Algorithm starts with a transformation matrix $W$ from MUSE, Procrustes or ICP and then refines it. The numbers 1, 2 and 3 represent runs over different seeds for non-deterministic methods (MUSE and ICP).}
\label{MUSE_table}
\end{table*}

We ran the Iterative Hungarian algorithm after normalizing the word embeddings (divide them by their Euclidean norm), which we found to converge faster. It must be noted that, since the adversarial part does not normalize the word embeddings, the $W_{0}$ matrices do not match exactly and thus not normalizing them should yield better results at a higher computational cost. \citet{NIPS2019_8836} showed that unit-length normalization makes GAN-based methods more unstable and also deteriorates their performance, but supervised alignments or Procrustes refinement are not affected by this.

The results can be seen in Table~\ref{MUSE_table}. 
We can see that our Iterative Hungarian algorithm improves the accuracy when used as a refinement tool. We believe that this is because the other methods do not try to optimize the Wasserstein-Procrustes objective directly, even though they achieve very good translations without relying on it. In the Appendix 
we report the performance of our algorithm on more language pairs. 

\rev{We also tried \citet{DBLP:journals/corr/abs-1906-01622}'s Iterative Normalization: before applying IH, we subtracted the mean of the word embeddings, and we normalized them. We repeated this process three times, and then we applied IH. The results appear in Table~\ref{MUSE_table2}: although this method improved the initialization produced by MUSE, better results were obtained by simply normalizing the word embeddings (as shown in Table~\ref{MUSE_table}).}

\begin{table*}[ht]
  \centering
  \begin{tabular}{llllllllllll}
    \toprule
    \multicolumn{1}{c}{Method} & \multicolumn{1}{c}{en-es} & \multicolumn{1}{c}{es-en} & \multicolumn{1}{c}{en-fr} & \multicolumn{1}{c}{fr-en}  & \multicolumn{1}{c}{en-it} & \multicolumn{1}{c}{it-en}  &  \multicolumn{1}{c}{en-ru} & \multicolumn{1}{c}{ru-en} & \multicolumn{1}{c}{mean} \\
    \cmidrule(r){1-1}
    \cmidrule(r){2-2}
    \cmidrule(r){3-3}
    \cmidrule(r){4-4}
    \cmidrule(r){5-5}
    \cmidrule(r){6-6}
    \cmidrule(r){7-7}
    \cmidrule(r){8-8}
    \cmidrule(r){9-9}
    \cmidrule(r){10-10}
  MUSE  & 
81.7 & 
83.5 & 
\textbf{82.5} & 
81.9 & 
77.5 & 
77.7 & 
\textbf{45.3} & 
61.0 & 
73.9
          \\
  MUSE + IH  & 
\textbf{82.3} &
\textbf{84.0} &
82.3 &
\textbf{82.5} &
\textbf{77.9} &
\textbf{77.9} &
44.9 &
\textbf{61.9} &
\textbf{74.2}  \\
 \bottomrule
  \end{tabular}
  \caption{The Iterative Hungarian (IH) Algorithm starts with a transformation matrix $W$ from MUSE, applies the iterative normalization from \citep{DBLP:journals/corr/abs-1906-01622} and then it refines the mapping.}
\label{MUSE_table2}
\end{table*}

\subsection{Aligning Word Embeddings from the Same Data} 

\label{toys}
The second set of experiments justify that the simple iterative procedure displayed in Algorithm~\ref{IH} works and we explain under what circumstances it can be relaxed or needs some help in the form of either supervision or a natural initialization matrix $W_{0}$. For the following controlled experiments, we set the initialization matrix to be the identity. We experiment with the following four approaches:
\begin{itemize}
    \item \textit{Hungarian.} Run the Hungarian algorithm for only one iteration, and then taking the permutation matrix $P$ as the map. 
    \item \textit{Cut Iterative Hungarian (CIH).} Run the Hungarian algorithm to update $Y \leftarrow PY$ and $X\leftarrow XW$ (see Algorithm~\ref{CIH}).
    \item \textit{Iterative Hungarian (IH).} Run the previous iterative procedure but considering the different natural initializations (see Algorithm~\ref{IH}).
    \item \textit{Supervised Iterative Hungarian (SIH).} Learn the correct mapping from a random 5\% subsample of the words, and then we run the IH algorithm for the remaining words. 
\end{itemize}
The experiments from this subsection recreate those by \citet{DBLP:conf/aistats/GraveJB19}; the idea is that English word embeddings are trained after changing some parameters, and the different spaces of word embeddings are rotated in order to match. We use fastText \citep{bojanowski2017enriching,joulin2017bag} to train word embeddings on 100M English tokens from the 2007 News Crawl corpus.\footnote{\url{http://statmt.org/wmt14/translation-task.html}} 

The different experiments in this section consist of changing the different training conditions and correctly mapping the results. We train the models using Skipgram \citep{article} unless stated otherwise, using the standard parameters of fastText.\footnote{\url{https://github.com/facebookresearch/fastText}} We perform four experiments: 
\begin{itemize}
    \item \textbf{Seed.} We only change the seed used to generate the word embeddings in our fastText runs. The source and the target are word embeddings trained using the same parameters.
    \item \textbf{Window.} We use window sizes of 2 and 10, respectively. The source and the target correspond to word embeddings trained on the same data but with different window sizes.
    \item \textbf{Algorithm.} We train the first algorithm with Skipgram and the second one with CBOW \citep{article}. The source and the target correspond to word embeddings trained on the same data but using a different method.
    \item \textbf{Data.} We separate the dataset in two different parts of the same length. We train corresponding word embeddings from the two separate parts. The source and the target correspond to word embeddings trained with the same parameters but on different data.
\end{itemize}

\begin{table}[t]
  \small
  \centering
  \begin{tabular}{lllll}
    \toprule
    \multicolumn{1}{c}{Method} & \multicolumn{1}{c}{Seed} & \multicolumn{1}{c}{Window} & \multicolumn{1}{c}{Algorithm} & \multicolumn{1}{c}{Data}  \\
    \cmidrule(r){1-1}
    \cmidrule(r){2-2}
    \cmidrule(r){3-3}
    \cmidrule(r){4-4}
    \cmidrule(r){5-5}
    Hungar. & \textbf{99\%}               &          7\%                     &       7\%                   &    1\%              \\
    CIH &   \textbf{100\%}     &      \textbf{100\%}     &               \textbf{100\%}               &       0\%                    \\
    IH &    \textbf{100\%}                &       \textbf{100\%}                    &            \textbf{100\%}                  &  0\%    \\ 
    SIH & \textbf{100\%}              &       \textbf{100\%}                      &              \textbf{100\%}                &          \textbf{100\%}                 \\
    \bottomrule
  \end{tabular}
\caption{Our method correctly aligns the word embeddings. \rev{\emph{Hungar.} is short for \emph{Hungarian}.}}
\label{toy}
\end{table}

We run the above algorithms on the 10,000 most frequent words. Table~\ref{toy} shows the results for the different algorithms. We perform the final mapping using the nearest neighbor for CSLS with $k = 10$, and the reported score is the percentage of words correctly mapped. Notice, that since we are \textit{translating English to English}, the correct map is trivial. Some observations follow:
\begin{itemize}
    \item The supervised approach works well with very little supervision, but all other attempts failed when facing the problem of mapping data from different datasets. This is probably because, by adding some supervision, we improve the initial $W_{0}$.
    This effect may be similar (although with less impact) to the help introduced in the IH algorithm with the equivalent problems or the natural initial transformations.
    \item The first three experiments converged in three iterations or less. The SIH algorithm took around twenty iterations to converge for the \emph{Data} experiment. 
    \item The Hungarian algorithm, which was not designed for the Wasserstein-Procrustes method, correctly finds the mapping for the seed experiment, whereas some other reported iterative experiments failed to achieve good results in this experiment \citep{DBLP:conf/aistats/GraveJB19}. 
\end{itemize} 
The proposed iterative procedures do converge, but they usually need good initial conditions or the help of supervision to converge to a good minimum. This suggests that Algorithm~\ref{CIH} could work well as long as we start from an initial transformation matrix $W_{0}$ close enough to the true solution.
The importance of the initial condition can be shown by the natural initial conditions. The solution of the four different equivalent problems induce different optimal transformation matrices $W^{*}$. In the first iteration of the IH algorithm, a branch among these four is chosen. Table~\ref{toy-dist} shows the Euclidean distance between each of the four natural initializations (assuming $W_{0} = \mathbb{I}$) and their respective optimal solution $W^{*}$ for the four experiments. These distances are different for the four branches, and to choose the best one (the one minimizing this distance) is key for convergence. 

\begin{table}[t]
  \small
  \centering
  \begin{tabular}{ccccc}
    \toprule
    \multicolumn{1}{c}{Method} & \multicolumn{1}{c}{Seed} & \multicolumn{1}{c}{Window} & \multicolumn{1}{c}{Algorithm} & \multicolumn{1}{c}{Data}  \\
    \cmidrule(r){1-1}
    \cmidrule(r){2-2}
    \cmidrule(r){3-3}
    \cmidrule(r){4-4}
    \cmidrule(r){5-5}
   $\mathbb{I}$ &    \textbf{9.49}                   &         \textbf{ 12.59}                   &  \textbf{12.45}                              &   \textbf{14.11}                  \\
   $V_{X}$ &   14.13            &      14.14                      &                        14.18        &                    14.19       \\
    $V_{Y}^{\top} $ &    14.15                      &      14.18                       &            14.18                    &    14.14                       \\
    $V_{X} V_{Y}^{\top} $ &   13.95                        &           14.10                  &                 14.09               &     14.16                      \\
    \bottomrule
  \end{tabular}
\caption{Distance between the natural initialization and the optimal solution for the four experiments.}
\label{toy-dist}
\end{table}

The distances that are too big do not converge to a good solution. For the \emph{Seed} experiment, such a small distance explains why a single iteration of the Hungarian algorithm was enough for a strong result. The Window and the Algorithm do not converge when running on a branch different from the first one---also the one that has the smallest distance---and when they run on the first branch, they converge in a few iterations. Hence, being able to provide a good initial transformation matrix $W_{0}$ and to correctly discriminate what the best branches are is essential for this approach.

In the Appendix we present further experiments on English to Spanish that test whether our method can be used without a good initialization, but with little supervision. We found that our method works well when little supervision is given.



\section{Conclusion and Future Work} \label{sec:conclusion}
We have underlined some mathematical properties of the Wasserstein-Procrustes problem and hence used the concept of the different natural initialization transformations in an iterative algorithm to achieve improved results for mapping word embeddings between different languages. In particular, we have shown that it is possible to use our algorithm as a refinement tool \rev{for UCAE} and we have demonstrated improved results after using the transformation of \citet{conneau2017word} as the initialization matrix $W_{0}$. 
We hope that our rethinking of the Wasserstein-Procrustes problem would enable further research and would eventually help develop better algorithms for aligning word embeddings across languages,
especially taking into account that most unsupervised approaches try to minimize loss functions different from Objective~\ref{WP}. 

In future work, we plan to study other loss functions. We are further interested to see how well the objectives in Table~\ref{formalism} correlate with CSLS.
Finally, we plan combinations with other existing methods. 

\section{Limitations}

While our work provides valuable insights and improvements for unsupervised cross-lingual alignment of embeddings, there are some limitations to consider:

\begin{itemize}
\item Our analysis primarily focuses on non-contextual unsupervised word embeddings. In future work, it is essential to extend this analysis to contextualized word embeddings, which are prevalent in modern NLP applications and offer additional challenges and opportunities for alignment.

\item Our study is more theoretical in nature, and the Wasserstein-Procrustes problem may not always hold true in practice due to factors such as noisy datasets or significant differences among languages. Despite these potential discrepancies, we believe our unified framework can inspire future research for improving word embeddings and contribute to more effective algorithms in aligning them across languages.
\end{itemize}

Overall, these limitations highlight potential avenues for further research and emphasize the importance of continued exploration in the field of unsupervised cross-lingual alignment of embeddings.

\section{Ethics Statement}

As researchers in the field of natural language processing, we recognize the importance of addressing ethical considerations in our work. In this study, we focused on unsupervised cross-lingual alignment of embeddings, with the aim of improving alignment techniques and fostering further research in this area. Below, we outline some of the ethical aspects that we have considered in this research:

\begin{itemize}
\item \textbf{Fairness and Bias:} We are aware that word embeddings can unintentionally capture and propagate biases present in the training data. By improving alignment techniques across languages, our work could potentially contribute to the mitigation of biases and the promotion of fairness in multilingual applications. However, we also acknowledge that our methods could inadvertently introduce or amplify biases. Future work should include thorough assessments of potential biases in the embeddings and the development of strategies to address them.

\item \textbf{Accessibility:} Our research aims to advance unsupervised cross-lingual alignment methods, which can contribute to the democratization of NLP technologies by enabling their application in low-resource languages with minimal data requirements.

\item \textbf{Privacy:} As our work is based on unsupervised word embeddings pretrained on large text corpora, it is crucial to ensure that the underlying data does not contain sensitive or personally identifiable information. We have made efforts to use publicly available and well-vetted datasets for our experiments and evaluations, minimizing potential privacy concerns.

\item \textbf{Impact:} The advancements in unsupervised cross-lingual alignment could lead to improved performance in various multilingual NLP tasks, such as machine translation, cross-lingual information retrieval, and sentiment analysis. While these improvements can have positive effects, it is essential to consider potential misuse of such technologies and remain vigilant against unintended consequences.
\end{itemize}

\section*{Acknowledgements}
This research was sponsored in part by the United States Air Force Research Laboratory and was accomplished under Cooperative Agreement Number FA8750-19-2-1000. This material is based upon work supported in part by the U.S. Army Research Office through the Institute for Soldier Nanotechnologies at MIT, under Collaborative Agreement Number W911NF-18-2-0048. The work was also supported by the Technical University of Catalonia (UPC), the CFIS program and Fundació Cellex. The views and the conclusions contained in this document are those of the authors and should not be interpreted as representing the official policies, either expressed or implied, of the United States Air Force, the U.S. Government, or the U.S. Army. The U.S. Government is authorized to reproduce and to distribute reprints for Government purposes notwithstanding any copyright notation herein.

\section*{References}
\bibliography{anthology,custom}
\bibliographystyle{lrec-coling2024-natbib}

\end{document}